\documentclass[12pt]{article}

\usepackage{graphicx}
\usepackage{amssymb}
\usepackage{url}
\usepackage{subfigure}
\usepackage{hyperref}
\usepackage{epstopdf}

\begin{document}

\title{Predicting Effective Control Parameters for Differential Evolution using Cluster Analysis of Objective Function Features\footnote{This is a pre-print of an article published in the Journal of Heuristics \href{https://doi.org/10.1007/s10732-019-09419-8}{DOI 10.1007/s10732-019-09419-8}. Cite this article as:
Walton, S.P. \& Brown, M.R. J Heuristics (2019). https://doi.org/10.1007/s10732-019-09419-8}}

\author{Sean P.~Walton \\ Department of Computer Science, College of Science \\ Swansea University, SA2 8PP, UK \\
              {s.p.walton@swansea.ac.uk}\\ \\ M.~Rowan Brown \\Centre for Nanohealth, College of Engineering \\ Swansea University, SA1 8EN, UK  
}

\date{\today}

\maketitle

\begin{abstract}
A methodology is introduced which uses three simple objective function features to predict effective control parameters for differential evolution. This is achieved using cluster analysis techniques to classify objective functions using these features. Information on prior performance of various control parameters for each classification is then used to determine which control parameters to use in future optimisations. Our approach is compared to state--of--the--art adaptive and non--adaptive techniques. Two accepted bench mark suites are used to compare performance and in all cases we show that the improvement resulting from our approach is statistically significant. The majority of the computational effort of this methodology is performed off--line, however even when taking into account the additional on--line cost our approach outperforms other adaptive techniques. We also study the key tuning parameters of our methodology, such as number of clusters, which further support the finding that the simple features selected are predictors of effective control parameters. The findings presented in this paper are significant because they show that simple to calculate features of objective functions can help to select control parameters for optimisation algorithms. This can have an immediate positive impact the application of these optimisation algorithms on real world problems where it is often difficult to select effective control parameters. 
\end{abstract}


\section{Introduction}
\label{}
Despite having a number of advantages over classical gradient based techniques, the performance of evolutionary algorithms depends both on the problem to be optimised and the algorithm being used~\cite{wolpert97}.  To make matters worse, this performance also depends heavily on the selection of algorithm specific control parameters.  This variability of performance makes the field hard to penetrate for users in industry who simply want to use an algorithm to solve a problem.  Often the problem they wish to solve is not well understood before they start to solve it, which makes selecting an algorithm and control parameters all the more difficult.  The motivation of the work, presented in this paper, is to automate this selection using simple machine learning techniques.  Specifically the aim is to automatically select an effective set of control parameters for differential evolution for an unknown problem.

\subsection{Terminology}
The problem to be optimised is termed the objective function. This paper focusses on optimising continuous black box objective functions. We identify a number of features, $\vec{\beta}$, that an objective function can be described by. An optimisation algorithm instance is determined by its control parameters $\vec{p}$. Our aim is to classify objective functions using their features, in order to predict a set of effective control parameters which will result in a high performing algorithm for a particular objective function. 

\subsection{Background}
When applying an evolutionary algorithm to a new application it is common to use the control parameters suggested in literature.  These parameters are usually obtained from extensive studies on algorithm behaviour using suites of benchmark optimisation problems.  Parameters which work well on common problem test suites will emerge~\cite{Eiben201119} and this single set will end up being used in the majority of applications.  The problem is that with truly novel applications there may be no understanding of which test suites, if any, correctly represents the real world problem.  Strictly speaking, each time an algorithm is applied to a new application a parameter study should be undertaken, to both provide insight into the robustness of the parameters and perhaps squeeze out some additional performance.  The reality is that these studies are often infeasible in real applications, where a single objective function evaluation may represent hours, or days, of computational time~\cite{david1015,lattice1,Walton2012,WaltonARCME}. Thus a great deal of research has been undertaken with the motivation to address this problem. We have identified three interrelated strands of research in the meta--heuristic optimisation community relevant to this problem. These are briefly discussed below, we then place our own approach in this context.

\subsubsection{Automatic Tuning Algorithms based on Performance Modelling}
A considerable body of work has shown that it is possible to build empirical performance models of algorithms~\cite{Hutter2014}. These models can then be used to select tuning parameters with good predicted performance~\cite{Hutter2006}. Sequential Model-Based Optimization for General Algorithm Configuration (SMAC)~\cite{HutHooLey11-smac} and Sequential Parameter Optimization (SPO)~\cite{bartz2005sequential} are both specific examples of this approach. In the case of SPO the approach facilitates manual tuning whereas SMAC is automated.

\subsubsection{Feature Based Approaches}
It is increasingly argued that we need to understand and use the characteristics and features of a problem to select a suitable algorithm, or tune it~\cite{SmithMiles2008}. Feature Based Algorithm Configuration (FBAC)~\cite{belkhir2016feature} can be thought of as an extension of the automatic tuning algorithms mentioned above. It uses sophisticated objective function features to classify objective functions. They are able to accurately predict performance models for objective functions which could, in theory, be used to determine an effective set of control parameters. However, the features they use require a large number of samples of the objective function to calculate. This would lead to an excessive computational cost in real applications. Exploratory landscape analysis (ELA)~\cite{Mersmann2011} introduces ten features, which are relatively cheap to calculate and can be used to classify objective functions. These features are grouped into five classes which relate to different characteristics of objective functions. Promising results have been presented whereby the ELA features are used to train a one--sided support vector regression model to select an appropriate optimisation algorithm~\cite{Kerschke2016}.

\subsubsection{Adaptive Algorithms}
The most common strategy to address the problem of performance variability is to design algorithms with self--adaptive control parameters.  In such algorithms the control parameters are themselves optimised, based on current performance, as the algorithm runs~\cite{Sarker2014,Zamuda201572,Guo201452}.  A related field is Hyper--Heuristics whose goal is to automate the design of heuristic optimisation algorithms based on current performance~\cite{Burke2013,game}. These strategies are performed on a per--objective function basis and do not use knowledge of objective function features, or past performance on different objective functions.

\subsubsection{Case Study Optimisation Algorithm: Differential Evolution}
\label{opt}
To show the effectiveness of our approach we are forced to select a single optimisation algorithm. Differential Evolution (DE)~\cite{storn97} will be used to test the effectiveness of the predictive methodology.  It is stressed that this approach is independent of the evolutionary algorithm, although some thought will be required if an algorithm has any non--continuous control parameters.  DE is popular and its control parameters are well studied. DE is aimed at nonlinear non--differentiable continuous functions and has been designed to be a direct stochastic search method. The method has a small number of control parameters and applies a crossover and mutation operator based on the differences between randomly selected individuals of the population.  

There are a number of alternative DE methods and many additions have been made to the algorithm.  It is beyond the scope of this paper to explain these additions in detail, so instead we describe the algorithm used in this study and allow the reader to find detailed explanation in original papers.

To select new members of the population, a direct one--to--one competition scheme is employed in each generation.  From the population of the current generation, a target member, ${\vec{x}}_{i,g}$, is selected, where $i$ refers to the member's number and $g$ the generation.  A donor vector, ${\vec{v}}_{i,g}$, is generated using the current-to-pbest/1/bin approach~\cite{Jingqiao_Zhang_2009}.  Three members of the population, distinct to that of the target member, are selected at random and  ${\vec{v}}_{i,g}$ is calculated according to the relation 
\begin{equation}
\label{deeq}
{\vec{v}}_{i,g} = {\vec{x}}_{i,g} + p_{2}({\vec{x}}_{pbest,g} - {\vec{x}}_{i,g}) + p_{2}({\vec{x}}_{r1,g} - {\vec{x}}_{r2,g})
\end{equation} 
where $p_{2}$ is a control parameter usually referred to as the weighting factor. ${\vec{x}}_{r1,g}$ and ${\vec{x}}_{r2,g}$ are two members selected at random from the whole population and ${\vec{x}}_{pbest,g}$ is randomly selected from the top $q \times p_{3} \: \:(q \in [0,1])$. $p_{3}$ is the population size or number of parents.  $q$ is a control parameter which controls the greediness of the algorithm, to eliminate this parameter it is randomised as in the Success--History Based Parameter Adaptation for Differential Evolution (SHADE) algorithm~\cite{Tanabe_2013}.  In addition, an external archive of previous members of the population is maintained and used to generate ${\vec{x}}_{r2,g}$~\cite{Tanabe_2013}.

A cross over operator is applied to the target and donor vectors to form a trial vector.  The elements of the target and donor vectors enter the trial vector with a probability $p_{1}$, a control parameter usually referred to as crossover constant. The target vector is compared with the trial vector and the vector with the best fitness value is selected for admission into the next generation. This iteration scheme repeats until a suitable stopping criterion is met~\cite{storn97}. 

DE has been applied, with success, to the fields of electrical power systems, electromagnetic engineering, control systems and robotics, chemical engineering, pattern recognition, artificial neural networks and signal processing~\cite{das2011}.  In~\cite{DEweb} Storn suggests using the control parameters $p_{1}=0.900$, $p_{2}=0.500$ and $p_{3}=10D$ where $D$ is the number of dimensions in the function.  The effect of these parameters on algorithm performance is a well researched subject.  For example, there appears to be complex relationships between problem dimensionality and the most appropriate population size~\cite{Piotrowski2016}. 

We compare the proposed predictive technique to a state of the art adaptive technique: SHADE~\cite{Tanabe_2013}.  This technique uses an historical memory of control parameters which have performed well to guide the selection of control parameters each generation.  In the original study it was shown to have competitive performance compared to other state of the art algorithms using the CEC 2005 benchmarks which are used in this study.  All control parameters used in our study are the same as used in the original SHADE study~\cite{Tanabe_2013}.

\subsection{Contribution and Motivation of this Paper}
The approach we have adopted is to select three simple to calculate features and use these to classify objective functions. Then as we optimise a series of objective functions a global memory of the performance, of various control parameters, for each of the classifications is stored. This information is then used to adapt control parameters for future optimisations. We do not create a performance model but directly use prior knowledge to adapt the optimisation algorithm. Thus our approach falls under the adaptive algorithm category and hence we compare our strategy to other adaptive strategies below. Our approach also falls under the feature based approach category since we are using objective function features to drive our adaptation. Our features are much simpler, and more crude, than those used in FBAC~\cite{belkhir2016feature} and we use fewer than those identified in ELA~\cite{Mersmann2011}. Our contribution is that even when using our deliberately simple approach there is a statistically significant improvement in performance when compared to algorithms which do not consider objective function features. The motivation for this is real world applications where it is infeasible to tune an algorithm each time a new objective function is considered, and where the form of the objective function may be unknown, making it difficult to relate to previous analyses of control parameters.

\section{Methodology}
\label{}

\subsection{Our Approach: Predicting Effective Control Parameters for Evolutionary Algorithms using Cluster Analysis of Objective Function Features}
The aim of our approach is to automatically predict an effective set of control parameters for an unknown objective function. This is achieved by classifying objective functions using three simple to calculate features which are described in Section~\ref{sec:features}. A number of experiments are performed off-line with varying control parameters, across a range of objective functions. The algorithm performance is measured and recorded for each experiment, the performance metric used is described in Section~\ref{sec:performance}. Functions are split into classifications using the unsupervised machine learning technique k-means++~\cite{Arthur2007}. All the experiments in a particular classification are ranked by performance and the mean values of the control parameters used in the top 10\% of experiments is calculated. When a new function is to be optimised it is sampled, on-line, and its features calculated. It is then classified and the mean values calculated for its classification are used to optimise it.

In the remainder of this section each aspect of this predictive approach is discussed in detail.

\subsubsection{Optimisation Algorithm Performance Metric}
\label{sec:performance}
There are a number of metrics which can be utilised to define the performance of an optimisation algorithm~\cite{Eiben201119,belkhir2016feature}.  The meaning of performance may change depending on the application~\cite{L_pez_Ib_ez_2016}, but in general we wish to reduce the objective function value with a small number of objective function evaluations.  In this work a performance metric, $\alpha$, is defined as
\begin{equation}
\alpha = \frac{100(F_{1}-F_{G})}{F_{1}N_{g}}
\end{equation}
where $F_{1}$ is the lowest objective function value in the first generation, $F_{G}$ is the lowest objective function value in the final generation, $G$, and $N_{g}$ is the total number of function evaluations performed up to and including generation $g$.  Generation $g$ is the first generation at which the reduction in the objective function reaches $99\%$ of the total reduction i.e.
\begin{equation}
\frac{F_{G}}{F_{g}} > 0.99
\end{equation}
This choice is justified as follows.  In practice, an evolutionary optimisation algorithm is run until a maximum number of objective function evaluations is reached or a predetermined accuracy, or tolerance, is achieved.  Dividing by $N_{g}$ means that $\alpha$ gives us information on the efficiency of the optimisation algorithm.  An algorithm which finds the optimum in the first few generations therefore has a larger $\alpha$ than an algorithm which found the optimum in the final generation.  In real applications, practicalities such as objective function evaluation cost, limit the number of objective function evaluations~\cite{david1015,lattice1,Walton2012,WaltonARCME}. $\alpha$ is designed to reward algorithms which exhibit high convergence in the first few function evaluations.  

It is not claimed that $\alpha$ is the correct metric for all situations, it is a choice depending on user requirements.  In this study an attempt is made to model a situation where an engineer wishes to apply an optimisation algorithm to a real problem.  One can imagine that such an engineer would simply select an algorithm and use the set of control parameters suggested in literature.  In the authors experience, in applying optimisation algorithms to engineering applications, the proposed metric $\alpha$ is relevant for many engineers.  Control parameters suggested in literature may not have been tuned with this metric in mind, despite this the engineer would likely use these parameters. In the results section convergence curves are presented to show the effect of this metric choice.  

\subsubsection{Objective Function Features}
\label{sec:features}
Functions are often described using features such as symmetry, smoothness, condition number or separability.  These are usually only defined for analytical problems.  It is well understood that these features affect the performance of optimisation algorithms.  The challenge, therefore, is to formulate a set of features that can be calculated with a small number of objective function evaluations.  

In this proof of concept study, the starting point for calculating these features will be a Latin hypercube sampling of the objective function search space. The number of samples taken is referred to as $\sigma$.  This sampling will be performed prior to the optimisation in this study, but in future this sampling could also be used as the first generation of the optimisation algorithm.  The objective function values in this sampling are first normalised by subtracting the mean and dividing by the standard deviation.  Three simple features have been selected to test the methodology.\begin{enumerate}
\item $\beta_{1}$, is the number of dimensions of the function which is known to strongly effect algorithm performance.
\item $\beta_{2}$, is the interquartile range of the normalised data, which provides information on function variation within the domain.  This feature will identify functions which have largely flat topology. This is identified as a feature which relates to curvature in~\cite{Mersmann2011}.
\item $\beta_{3}$, is the skew of the normalised data.  The skew tells us how the function value is distributed, a skew of zero would indicate a normal distribution, whereas positive and negative values would indicate a tailed distribution.  This feature could potentially identify functions with sharp optimum as well as give information regarding function symmetry. This is identified as a feature which relates to y--distribution in~\cite{Mersmann2011}.
\end{enumerate} Collectively these features make up the characteristics, $\vec{\beta}$, of a particular objective function.

\subsubsection{Control Parameter Selection}
DE requires a number of control parameters, stored in the vector $\vec{p}$, which defines a single instance of the algorithm.  Running many $\vec{p}$ on many objective functions results in a number of data points in the form $(\vec{p},\vec{\beta},\alpha)$.  This data is named the training data and is used to exploit any relationships between the control parameters, function features and performance.  

The approach adopted is to apply the unsupervised clustering algorithm k-means++~\cite{Arthur2007}.  The k-means algorithm takes an unlabelled data set and classifies it into a user specified number of groups, $\kappa$.  Each group is defined by a cluster centroid, a data point belongs to the group whose centroid it lies closest to.  The k-means++ variant of the algorithm carefully initialises these centroids in favour of random initialisation~\cite{Arthur2007}.  

Using the training set, the objective functions are classified by applying k-means++ to the $\vec{\beta}$ data points.  For each classification the data points are sorted by $\alpha$.  The top $10\%$ data points are identified and the mean $\vec{p}$ is calculated from that set and used to optimise new functions identified as belonging to that classification.  

At the end of optimisation the new data point, $(\vec{p},\vec{\beta},\alpha)$, from that run is appended to the training data and the k-means++ algorithm is run to update the classifications and redetermine the best control parameters for each new centroid.  The key idea is that the memory of good performing parameters are extended from a single optimisation run to the entire history of using the algorithm.

\subsection{Experimental Methodology}
\label{test}
\subsubsection{Procedure for a Single Optimisation}
Each time a function was optimised the optimiser was limited to $10,000$ objective function evaluations. To consider our approach fairly the on-line cost of calculating $\vec{\beta}$, before the optimisation takes place, contributes to the number of objective function evaluations. In other words our approach has fewer objective function evaluations available when the optimisation starts. All optimisation runs were repeated 30 times, with 30 different random seeds used for all random number generation.  With the same random seed the same control parameters would result in the same performance on the same function instance. This allowed pairwise comparisons between different control parameters.  Repeating each optimisation run with 30 different random seeds ensured a 'lucky' seed was not selected which benefited a particular approach.

\subsubsection{Test Suites}
Two established optimisation benchmark suites were used in this study.  

\paragraph{Real--Parameter Black--Box Optimization Benchmarking functions (BBOB) 2015}
BBOB 2015 were used to train the predictive methodology.  The 24 benchmark functions which make up the BBOB 2015 test suite are given in~\cite{wp200901_2010,hansen2010fun}.  This suite includes separable functions, functions with low to high condition numbers and multi-modal functions with weak global structure.  The same numbering system for the functions in~\cite{hansen2010fun} is used in this paper.  All of these functions are defined for an arbitrary number of dimensions and have the same search domain.  The test suite includes $15$ instances for each function, for each instance a combination of optimal location shifting and linear transformations are applied.  Each instance is shifted and rotated in the same manor on subsequent runs which enables direct comparison of performance.  In the experiments presented here, a single test suite entails optimising each function instance at a range of dimensions (2, 10, 20, 30, 40, 50) using 30 different random seeds.  The resulting number of tests in a single run of the suite is then $64,800$.   

\paragraph{IEEE Congress on Evolutionary Computation CEC 2005 Real-Parameter Optimisation benchmarks}  
The CEC 2005 benchmark functions as detailed in~\cite{CEC2005} make up the second test suite.  These were used to test the effectiveness of problem specific tuning on objective functions different to the training set. The 25 functions were used with the same numbering system presented in the technical report~\cite{CEC2005}.  All functions were optimised at 2, 10, 30 and 50 dimensions using 30 different random seeds resulting in a total of $3,000$ tests.

\subsubsection{Statistical Methodology}
Using the test suites described above allows pairwise comparison of $\alpha$ between different approaches.  The approach for using nonparametric statistical tests described by Derrac et al.~\cite{Derrac20113} is followed here.  The Wilcoxon signed ranks test is used to compare the predictive methodology to other approaches.  The test results in the value $W$, the sum of the ranks of the differences (zero-differences are split between positive and negative) which will be reported along with the two sided \emph{p}-value. For all the statistics presented a positive $W$ indicates that the predictive methodology has performed better than the group it is compared to, a larger $W$ indicates a more significant improvement.   

\subsubsection{Experiments}
For each function in the BBOB 2015 suite a Latin hypercube sampling of $\vec{p}$ will be generated and each of these control parameter sets used to optimise that function.  Since there are three control parameters, 30 sets of $\vec{p}$ will be generated each time.  For these optimisation runs the number of samples used to calculate $\vec{\beta}$, was set to $\sigma=1000$.  The resulting data will be used as the initial training set for the problem aware tuning.  

There will then be four methods for selecting the DE control parameters:
\begin{itemize}
\item the suggested parameters from literature,
\item SHADE,
\item the predictive methodology (using cluster analysis),
\item using the best performing control parameters from the training set.
\end{itemize}
The predictive methodology will be applied with varying $\sigma$ and $\kappa$ to gauge the sensitivity to these. Each time our approach is used, a new set of samples is generated to calculate $\vec{\beta}$ in order to simulate the use of the approach in practice.  Each of these methods will be used to optimise both function suites.  The non--parametric tests will then be utilised to compare the effectiveness of each method.  

It needs to be stressed that, in the comparisons presented, $\vec{\beta}$ is recalculated and used for objective function classification in each optimisation run.  This does mean that when comparing the number of function evaluations to other methods, the predictive methodology has $\sigma$ additional evaluations.  These function evaluations have been included in all measurements of performance as they indicate the cost of our methodology.  

The goal of this paper is to show that features, such as those in $\vec{\beta}$, can be used as predictors for $\vec{p}$ in order to maximise $\alpha$.  If this is the case, future research into minimising the required $\sigma$ to effectively approximate $\vec{\beta}$ can be undertaken, as well as research into different definitions for $\alpha$.

\section{Results}
\label{}

\subsection{BBOB 2015 function suite}
\begin{figure*}[!t]
\centering
\includegraphics[width=\textwidth]{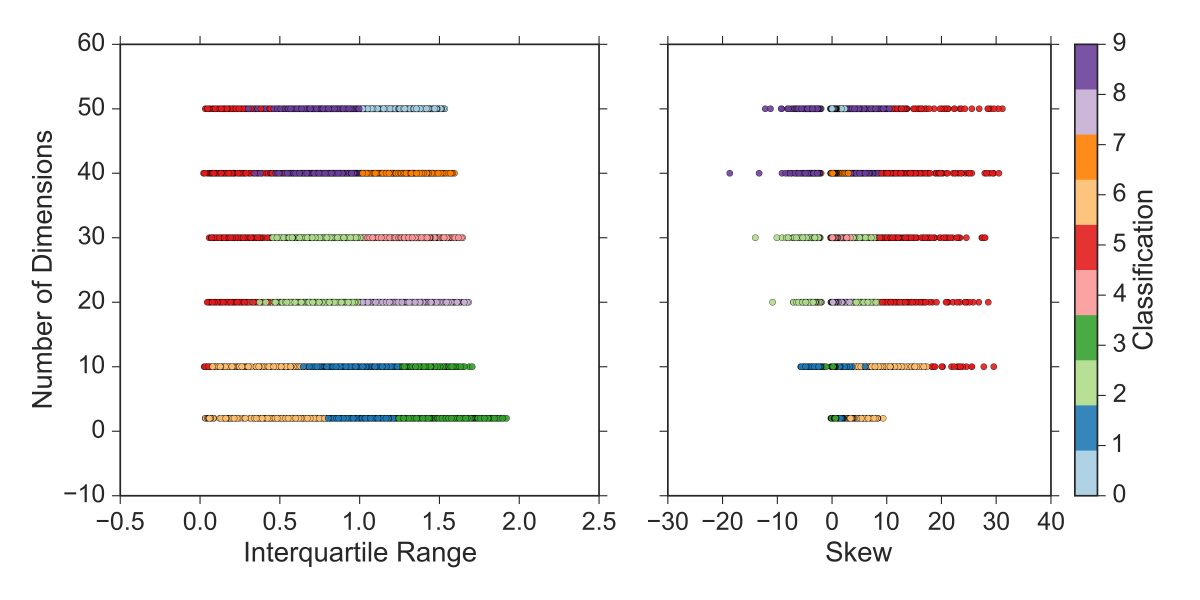}
\caption{The BBOB 2015 training data set.  Markers are coloured according to their classification when $\kappa=10$.}\label{BBOBTrain}
\end{figure*}
Fig.~\ref{BBOBTrain} shows a two dimensional projection of the training data used in the following studies.  This data resulted from optimising the BBOB 2015 function suite only.  The marker colour indicates which classification each data point belongs to when $\kappa=10$.   

In the following experiments the BBOB 2015 suite is both the training suite and the testing suite. This is the most basic test for our approach. It is worth pointing out that $\vec{\beta}$ is recalculated using new random samplings for each optimisation experiment. 

\subsubsection{Predictive Methodology Compared to Picking the Best from the Training Set}
\begin{table}[!t]
\renewcommand{\arraystretch}{1.3}
\caption{The control parameters used in the fixed control parameter optimisations}
\centering
\begin{tabular}{rlll}
\hline
\multicolumn{1}{l}{} & $p_{1}$ & $p_{2}$ & $p_{3}$  \\
\hline
Best in training set &   0.624  &  0.744   &   336      \\
Suggested in literature~\cite{DEweb}   & 0.900    & 0.500     & $10D$         \\
\hline
\end{tabular}
\label{fixed}
\end{table}
In the following study a single set of tuning parameters, the best performing in the initial training set, were used to optimise the BBOB 2015 test suite.  The control parameters used in this study are presented in Table~\ref{fixed}.  The results of the statistical tests, shown in Table~\ref{best}, show that predictive methodology results in a statistically significant increase of $\alpha$ compared to the best from the training set.  This improvement was achieved regardless of the values of $\kappa$ and $\sigma$.
\begin{table}[!t]
\renewcommand{\arraystretch}{1.3}
\centering
\caption{BBOB 2015 Function Suite: Wilcoxon signed ranks test data comparing the predictive methodology to using the best performing control parameters overall in the training set (equivalent to $\kappa=1$)}
\begin{tabular}{lllll}
\hline
$\kappa$ & $\sigma$ & $W$      & \emph{p}-value          \\ 
\hline
10       & 10       & 12647560.5    & \textless0.0001         \\
10       & 100      & 12000441.0    & \textless0.0001          \\
10       & 1000     & 14409261.5    & \textless0.0001           \\
100      & 1000     & 13138542.0    & \textless0.0001          \\ \hline
\end{tabular}
\label{best}
\end{table}

\subsubsection{Predictive Methodology Compared to Using the Best Parameters from Literature}
\begin{table}[!t]
\renewcommand{\arraystretch}{1.3}
\centering
\caption{BBOB 2015 Function Suite: Wilcoxon signed ranks test data comparing the predictive methodology to using the control parameters most commonly used in literature}
\begin{tabular}{lllll}
\hline
$\kappa$ & $\sigma$ & $W$      & \emph{p}-value          \\ 
\hline
10       & 10       & 64554898.0     & \textless0.0001          \\
10       & 100      & 65968426.5     & \textless0.0001           \\
10       & 1000     & 87498052.0     & \textless0.0001           \\
100      & 1000     & 83669358.0     & \textless0.0001            \\ \hline
\end{tabular}
\label{literature}
\end{table}
The test suite was optimised using the control parameters suggested in literature, these parameters are shown in Table~\ref{fixed}.  These parameters are what most practitioners would use in practice as a \emph{rule of thumb}.  Table~\ref{literature} shows the statistical comparison between the performance of these fixed parameters to the predictive methodology.  In all cases the predictive methodology performs significantly better.  There is a jump in performance when $\sigma$ increases from $100$ to $1000$ which indicates a sensitivity on the sampling of the objective functions.

\subsubsection{Predictive Methodology Compared to SHADE}
\begin{table}[!t]
\renewcommand{\arraystretch}{1.3}
\centering
\caption{BBOB 2015 Function Suite: Wilcoxon signed ranks test data comparing the predictive methodology to SHADE}
\begin{tabular}{lllll}
\hline
$\kappa$ & $\sigma$ & $W$      & \emph{p}-value          \\ 
\hline
10       & 10       & 50395288.5    & \textless0.0001         \\
10       & 100      & 51910557.5    & \textless0.0001          \\
10       & 1000     & 68985487.5    & \textless0.0001          \\
100      & 1000     & 66556975.0    & \textless0.0001           \\ \hline
\end{tabular}
\label{jDE1}
\end{table}
The test suite was optimised using SHADE.  Table~\ref{jDE1} shows the statistical comparison between the performance of SHADE to the predictive methodology.  In all cases the predictive methodology performs significantly better than adaptive tuning.  The performance is more significant when $\sigma=1000$.

\subsubsection{Convergence Behaviour}

\begin{figure}[!t]
\centering
\begin{subfigure}
	\centering
	\includegraphics[width=\textwidth]{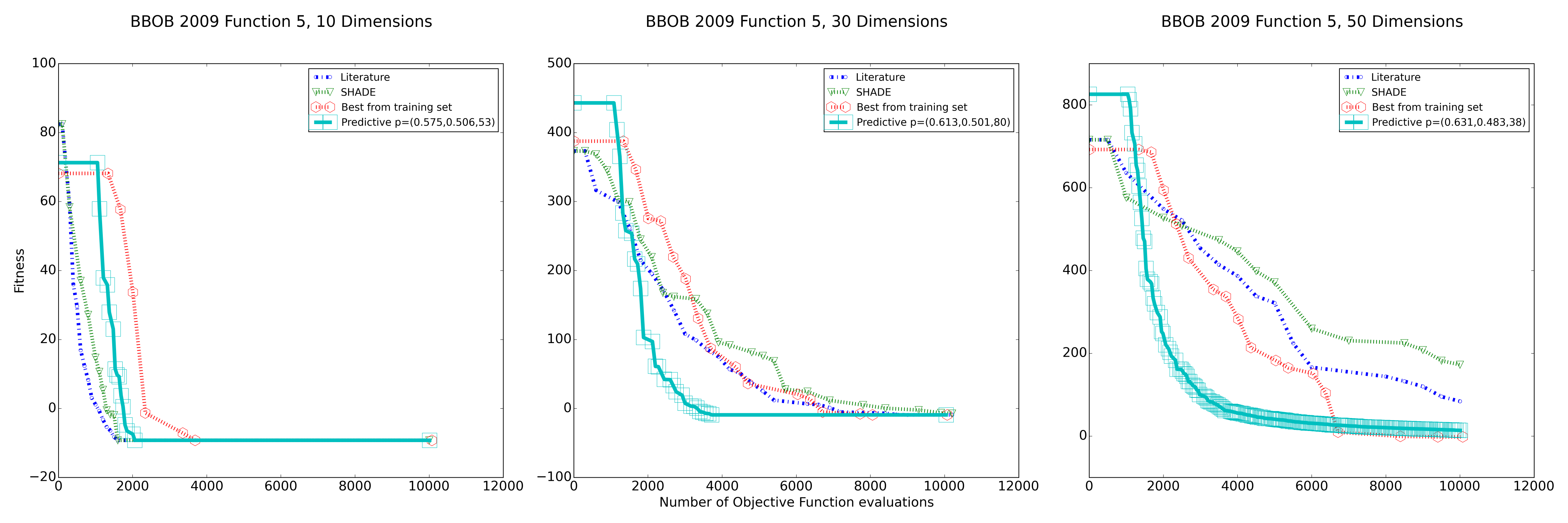}
\end{subfigure}
\begin{subfigure}
	\centering
	\includegraphics[width=\textwidth]{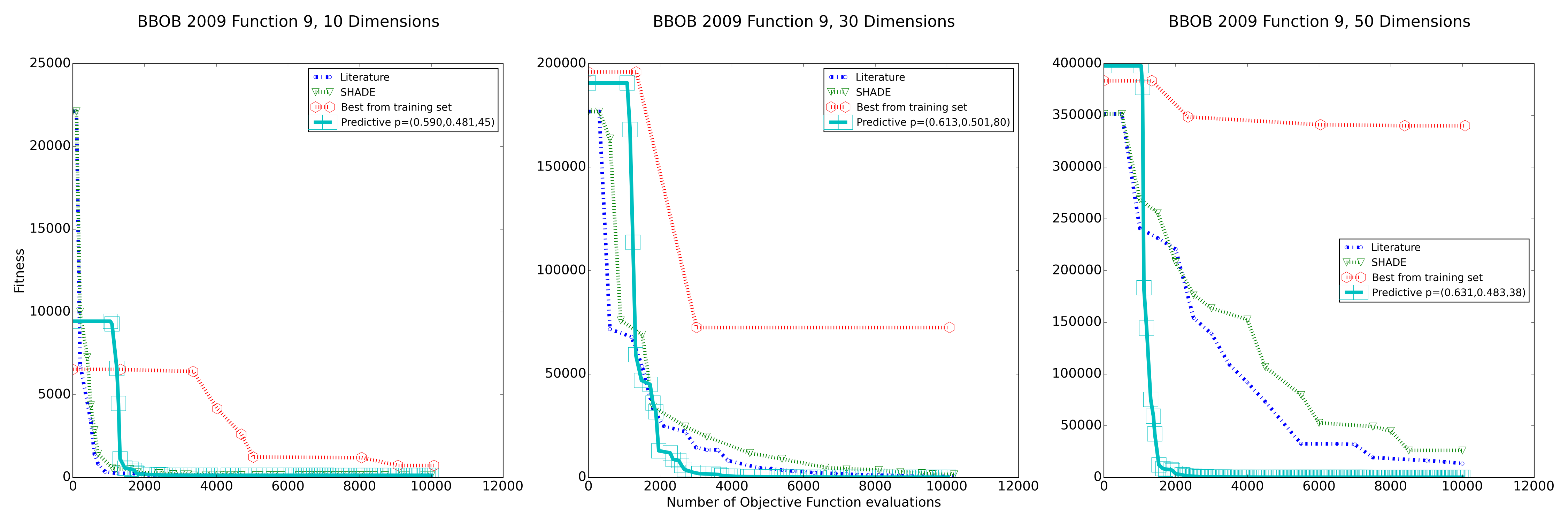}
\end{subfigure}
\begin{subfigure}
	\centering
	\includegraphics[width=\textwidth]{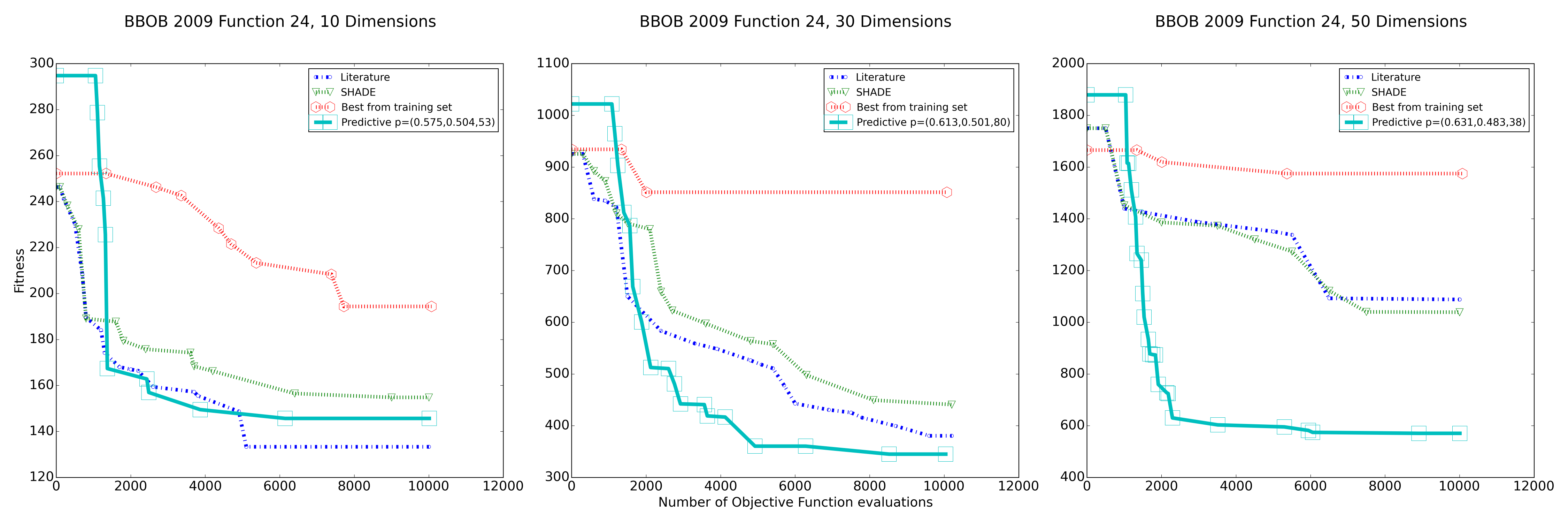}
\end{subfigure}
\caption{Examples of the convergence behaviour using different control parameter selection strategies for the BBOB 2015 function suite.  In the predictive methodology $\kappa=10$ and $\sigma=1000$.}\label{BBOBFuns1}
\end{figure}

In Fig.~\ref{BBOBFuns1} convergence plots are presented for a number of functions in the BBOB 2015 suite.  The objective function value is plotted against the number of function evaluations for each control parameter selection strategy.  These data points were only recorded once per generation if an improvement in the objective function value was found.  This means that the number of data points depends on the population size and is not the same for every curve.  Each function is shown for different numbers of dimensions and the control parameters selected by the predictive methodology are presented.  These functions were selected to show a range of cases, some where the predictive methodology performs well and some where it does not. Where the predictive methodology performs well it achieves rapid convergence early in the optimisation, which is what the metric $\alpha$ was designed to achieve.

\subsection{CEC 2005 function suite (predictive methodology trained using BBOB 2015)}
\subsubsection{Problem Aware Tuning Compared to Picking the Best from the Training Set}
\begin{table}[!t]
\renewcommand{\arraystretch}{1.3}
\caption{CEC 2005 Function Suite: Wilcoxon signed ranks test data comparing the predictive methodology to using the best performing control parameters overall in the training set (equivalent to $\kappa=1$)}
\centering
\begin{tabular}{lllll}
\hline
$\kappa$ & $\sigma$ & $W$      & \emph{p}-value       \\ 
\hline
10       & 10       & 55436.0    & \textless0.0001        \\
10       & 100      & 49811.5    & \textless0.0001           \\
10       & 1000     & 58615.0    & \textless0.0001           \\
100      & 1000     & 58971.5    & \textless0.0001           \\ \hline
\end{tabular}
\label{bestCEC}
\end{table}
The predictive methodology and the best control parameters from the initial training set were used to optimise the CEC 2005 benchmark functions.    Table~\ref{bestCEC} shows the statistical tests for this comparison.  For all $\kappa$ and $\sigma$ the predictive methodology performs significantly better with \emph{p}-values \textless0.0001.  The improvement observed when optimising the CEC 2005 function suite is comparatively less significant than the improvement observed when optimising the BBOB 2015 suite.  

\subsubsection{Predictive Methodology Compared to Using the Best Parameters from Literature}
\begin{table}[!t]
\renewcommand{\arraystretch}{1.3}
\centering
\caption{CEC 2005 Function Suite: Wilcoxon signed ranks test data comparing the predictive methodology to using the control parameters most commonly used in literature}
\begin{tabular}{lllll}
\hline
$\kappa$ & $\sigma$ & $W$   & \emph{p}-value         \\ 
\hline
10       & 10       & 220650.0    & 0.001         \\
10       & 100      & 223100.0    & 0.003           \\
10       & 1000     & 233488.0    & 0.075          \\
100      & 1000     & 243996.0    & 0.528          \\ \hline
\end{tabular}
\label{literatureCEC}
\end{table}
The results of statistical tests comparing the predictive methodology to the fixed parameters suggested in literature are shown in Table~\ref{literatureCEC}. For all cases the predictive methodology outperformed the fixed parameters, but the increase in performance is not statistically significant when $\kappa=100$. 

\subsubsection{The Predictive Methodology Compared to SHADE}
\begin{table}[!t]
\renewcommand{\arraystretch}{1.3}
\centering
\caption{CEC 2005 Function Suite: Wilcoxon signed ranks test data comparing the predictive methodology to using SHADE}
\begin{tabular}{lllll}
\hline
$\kappa$ & $\sigma$ & $W$     & \emph{p}-value        \\ 
\hline
10       & 10       & 183736.0      & \textless0.0001         \\
10       & 100      & 187102.0    & \textless0.0001          \\
10       & 1000     & 217967.0    & \textless0.0001          \\
100      & 1000     & 206840.0    & \textless0.0001           \\ \hline
\end{tabular}
\label{jDECEC}
\end{table}
The results of statistical tests comparing the predictive methodology to the fixed parameters suggested in literature are shown in Table~\ref{jDECEC}. For all $\kappa$ and $\sigma$ the predictive methodology significantly outperformed SHADE with \emph{p}-values all \textless0.0001.

\subsubsection{Convergence Behaviour}
\begin{figure}[!t]
\centering
\begin{subfigure}
	\centering
	\includegraphics[width=\textwidth]{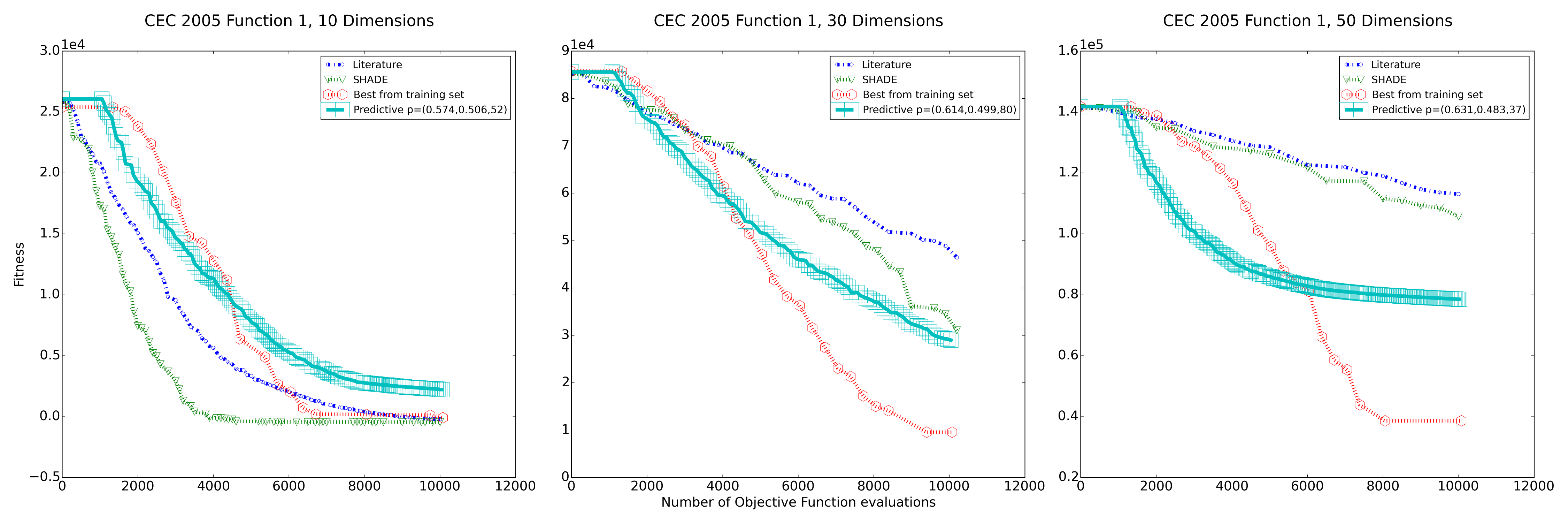}
\end{subfigure}
\begin{subfigure}
	\centering
	\includegraphics[width=\textwidth]{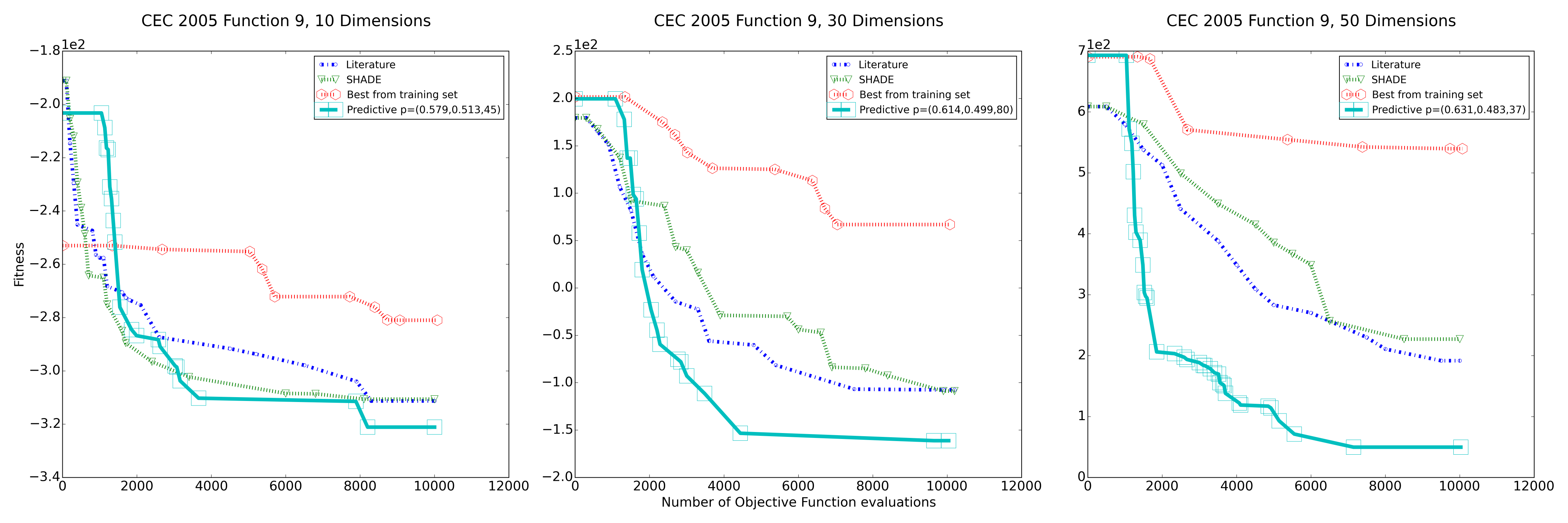}
\end{subfigure}
\begin{subfigure}
	\centering
	\includegraphics[width=\textwidth]{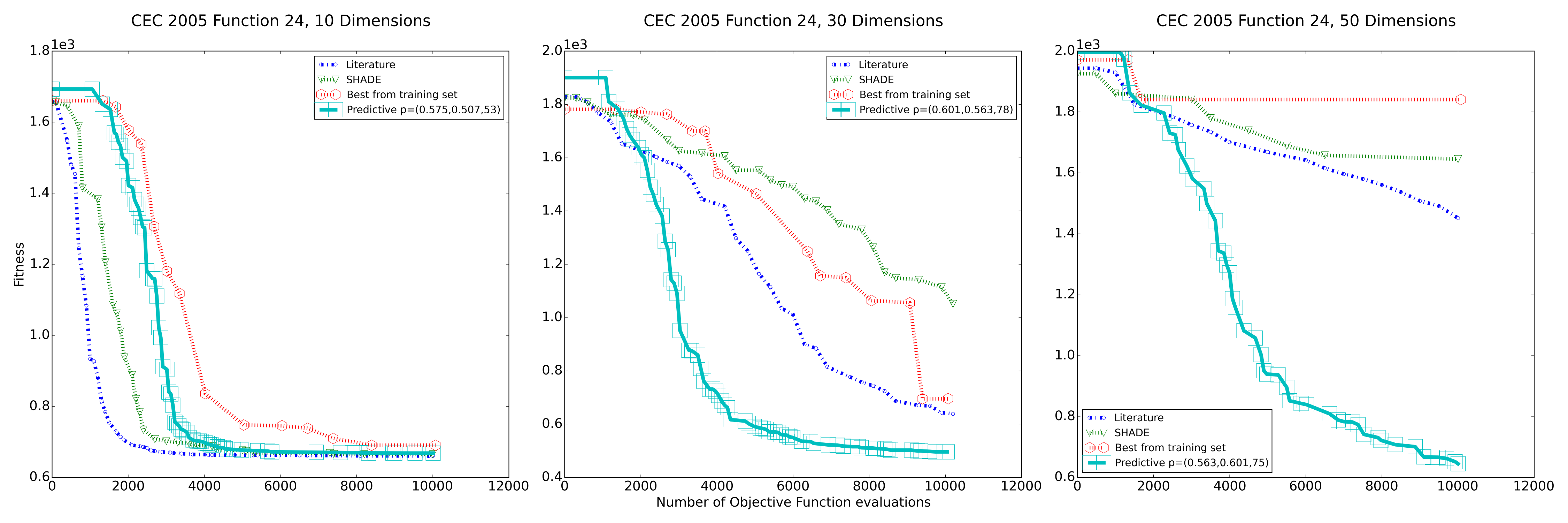}
\end{subfigure}
\caption{Examples of the convergence behaviour using different control parameter selection strategies for the CEC 2005 function suite. The predictive methodology was trained using the BBOB 2015 function suite with $\kappa=10$ and $\sigma=1000$.}\label{CECFuns1}
\end{figure}
Fig.~\ref{CECFuns1} compares convergence plots for functions from the CEC 2005 suite using the different control parameter selection strategies.  Functions were selected to present a range of behaviours.  

\section{Discussion}
The results show that, when optimising the BBOB 2015 objective function suite, the predictive methodology out performed using fixed and adaptive tuning parameters for DE with \emph{p}-values\textless0.0001.  The predictive methodology is more likely to outperform other methodologies when the initial sampling size, $\sigma$, of the objective functions was increased. There was a slight drop in performance from $\kappa=10$ to $\kappa=100$.  This indicates that having fewer large classifications of objective function are better than a more granulated approach.  This trend does not continue to $\kappa=1$, i.e. simply using the best parameters from the training set.  The observed improvement from $\kappa=1$ to $\kappa=10$ and $\sigma=100$ to $\sigma=1000$ both support the claim that $\vec{\beta}$ can be used as a predictor for which control parameters to use in an optimisation algorithm.  

Optimising the CEC 2005 benchmark functions, using only the BBOB 2015 suite for training, is a tough test for the suitability of $\vec{\beta}$ to act as a predictor for control parameters in the general case. In all cases the predictive methodology outperformed other approaches. The closest performing approach to the predictive methodology was simply using fixed control parameters suggested in literature, in this case two tests ($\kappa=10$, $\sigma=1000$ and $\kappa=100$, $\sigma=1000$) have \emph{p}-values $0.075$ and $0.528$ respectively. This could be explained by the fixed control parameters not requiring objective function evaluations to adapt, i.e. all objective function evaluations are used for optimisation. The high value of $\sigma$ for these outliers further supports this explanation, when $\sigma<100$ the \emph{p}-values are \textless0.01 when comparing to fixed control parameters from literature. This effect is then compounded when $\kappa=100$ which, as observed elsewhere, performs less well than $\kappa=10$.

Overall the results show that the simple objective function features, $\vec{\beta}$ can act as a predictor for selecting appropriate control parameters for DE.  The advantage of this approached can be observed when comparing it to SHADE.  SHADE learns which control parameters are most effective during the optimisation process.  The predictive methodology attempts to predict effective control parameters prior to the optimisation, therefore the benefit is felt from the first iteration.  This prediction itself comes at the cost of objective function evaluations, which were accounted for in this study. The open problem, therefore, is how to effectively approximate these features with fewer objective function evaluations.  In the future, an aim is to use these objective function evaluations as the first generation of the optimisation run in order not to waste them.

To improve performance, it may be possible to update the value of $\vec{\beta}$ as the optimisation runs and better samples the objective function.  As the approximation of $\vec{\beta}$ improves the control parameters could be changed mid run.  This would require thoughtful implementation to avoid introducing significant computational overhead.  There is also scope for designing more sophisticated and varied objective function features.  Performance may also be increased with a larger training data set. This does come with an additional overhead, as the computational cost of the k-means algorithm increases with the size of the training data.  In the future there is no reason why the k-means calculation could not be performed using cloud computing with training data collected from many users of the algorithm.

\subsection{Conclusions}
\label{}
The methodology proposed has been shown to offer statistically significant improvement over other approaches.  This implementation shows that the concept has the potential to be a powerful addition to evolutionary optimisation algorithms.  The method is general and could be applied to any evolutionary algorithm and any performance measure of interest.  There are a number of avenues to investigate, discussed above, which may improve the methodology further.  In particular an investigation into more sophisticated function features, such as those presented in ELA~\cite{Mersmann2011}. The long term goal should be to extend this methodology to automatically select the most appropriate evolutionary algorithm for a problem, not just the control parameters.  Such an automation would be of great use for industrialists wishing to apply evolutionary algorithms to real world applications.

\section{Acknowledgements}
The Welsh Government is acknowledged for a S\^{e}r Cymru II Fellowship [80761-SU-006] (A.O.W) part funded by the European Regional Development Fund (ERDF).


\end{document}